\documentclass[11pt]{article}

\usepackage[final]{acl}

\usepackage{times}
\usepackage{latexsym}
\usepackage{mathrsfs}
\usepackage{siunitx}
\usepackage{makecell}
\usepackage{booktabs}
\usepackage{placeins}
\usepackage[T1]{fontenc}
\usepackage{subcaption}

\usepackage[utf8]{inputenc}

\usepackage{microtype}

\usepackage{inconsolata}

\usepackage{graphicx}
\usepackage{amsmath,amssymb,amsfonts}
\title{SSPO: Subsentence-level Policy Optimization}

\makeatletter
\renewcommand{\@fnsymbol}[1]{*}
\makeatother

\author{
  Kun Yang$^{1}$\thanks{Equal contribution.} \quad
  Zikang Chen$^{1}$\footnotemark[1] \quad
  Yanmeng Wang$^{1}$ \\
  Zhigen Li$^{1,2}$ \quad
  Ning Cheng$^{1}$ \quad
  Shaojun Wang$^{1}$ \quad
  Jing Xiao$^{1}$ \\
  $^{1}$Ping An Technology $^{2}$TJUNLP Lab, Tianjin University\\
  \texttt{\{yangkun219, chenzikang003\}@pingan.com.cn}
}

\begin{document}
\maketitle
\begin{abstract}
As a key component of large language model (LLM) post-training, Reinforcement Learning from Verifiable Rewards (RLVR) has substantially improved reasoning performance. However, existing RLVR algorithms exhibit distinct stability issues: GRPO (Group Relative Policy Optimization) often suffers from unstable policy updates, while GSPO (Group Sequence Policy Optimization) can retain high-variance tokens. In GRPO, the importance ratio is computed at the token level, which overemphasizes individual tokens and makes learning sensitive to outliers, potentially causing training collapse. GSPO instead computes a response-level importance ratio, mitigating variance and reducing the accumulation of token-level noise present in GRPO. Nevertheless, our experiments show that GSPO frequently yields a near-zero clipping fraction: extreme token-level ratios can be diluted by other tokens in the same response, causing the entire response to be retained and resulting in unstable updates. We propose SSPO, which computes importance ratios at the subsentence level, striking a balance between GRPO and GSPO. SSPO alleviates training collapse and excessive variance while avoiding the failure mode in which the clipping mechanism indiscriminately retains entire responses. Moreover, we incorporate subsentence-level entropy into PPO-CLIP to adaptively adjust the clipping bounds: we encourage exploration for high-entropy tokens while tightening the clipping range for low-entropy tokens. Empirically, SSPO achieves an average score of 46.72 across five datasets on Qwen2.5-1.5B-Math model, outperforming GRPO (43.01) and GSPO (44.42), and attains state-of-the-art results on four datasets. On Qwen2.5-7B-Math model, SSPO also achieves the highest averaged scores over five baseline methods. These results demonstrate SSPO’s effectiveness in RLVR.
\end{abstract}

\begin{figure*}[t]
    \centering
    \includegraphics[width=1.0\textwidth]{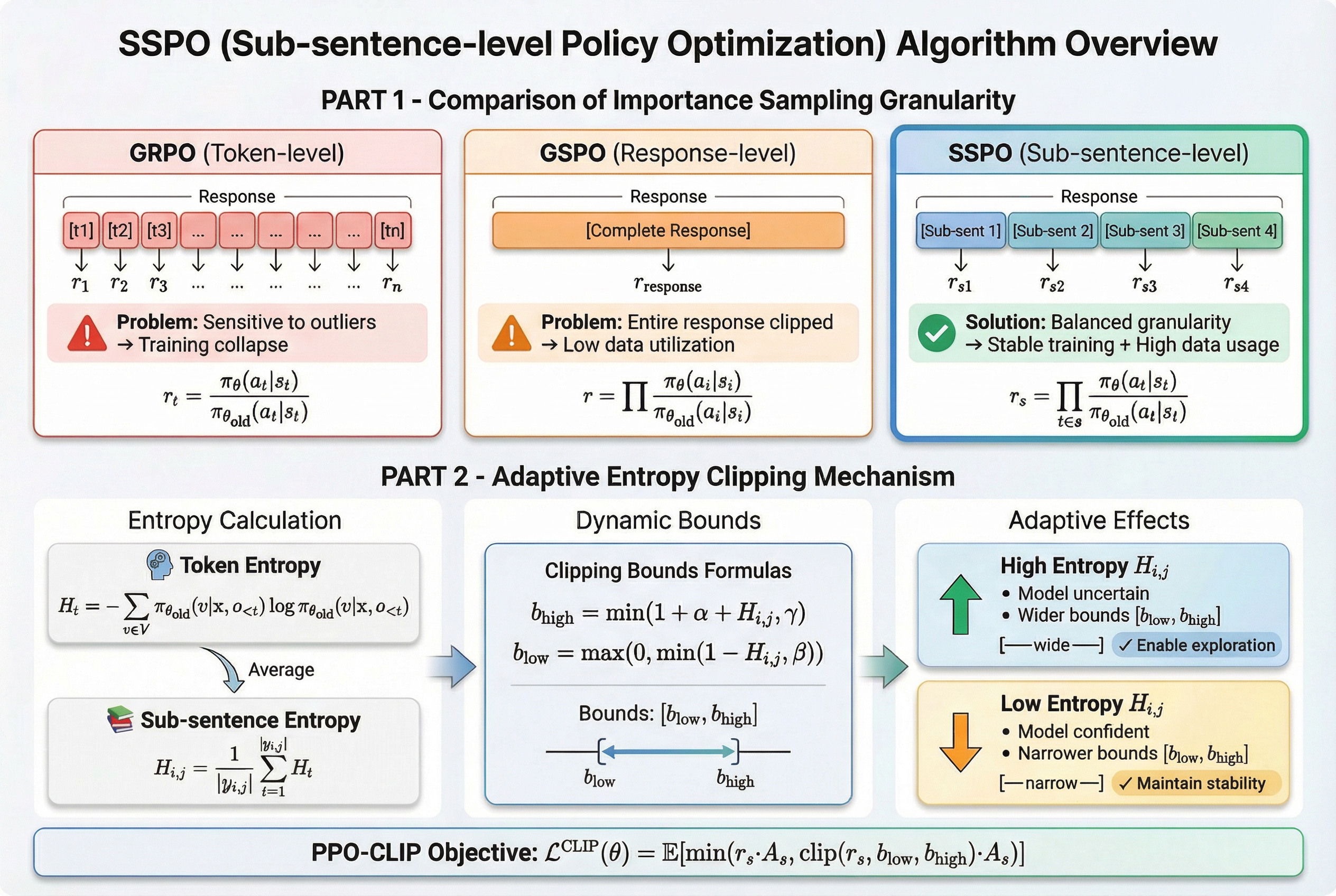}
    \caption{The overview of our proposed SSPO}
    \label{fig1}
\end{figure*}

\section{Introduction}
In recent years, Reinforcement Learning has greatly improved the reasoning capabilities of large language models (LLMs). Given verifiable rewards, language models can solve sophisticated problems, such as mathematics and programming, through large-scale RL. These RL methods allow LLMs to explore and apply a long chain-of-thought to tackle difficult problems. With increasing response length, the reasoning performance of LLMs has been markedly enhanced.
GRPO \citep{shao2024deepseekmath}, the main algorithm of RLVR, exhibits stability issues while training large language models, often resulting in high-variance training noise, leading to model collapse. In order to address these challenges, GSPO \citep{zhao2025geometric}, an excellent RL algorithm modified by GRPO,  introduces a response-level importance ratio and becomes more efficient and stable compared to GRPO. 

However, applying a response-level importance sampling ratio to RLVR training might cause other problems. Firstly, due to all the response tokens sharing a common importance sampling ratio, which has been calculated based on the average importance weight of each token in response, these tokens might all be clipped by the PPO-clip mechanism when a few tokens possess extremely high or low values of importance ratio. This will result in a high clipping rate and mass waste of the sampled data. Secondly, when there are a few abnormal values of the importance weight of tokens in a response, these anomalies might be preserved in the gradient after taking the average. These preserved tokens possibly affect the training stability and training efficiency.

In this paper, we present SSPO, a novel RL algorithm designed to address the core shortcomings caused by GSPO. First of all, we introduce a subsentence-level importance weight by splitting the whole response into a few sentences. Each token in a sentence has one importance sampling weight. In this way,  it can prevent the entire response from being abandoned by PPO-clip, and reduce the clipping fractions. Also, we introduce entropy-adaptive clipping to dynamically adjust the clipping bounds depending on the sentence entropy, encouraging high-entropy tokens to explore further and limiting the clipping range of low-entropy tokens. In this way, entropy collapse has been effectively avoided and the training stability of RLVR has been significantly improved.
Finally, we used levels 3-5 of MATH \citep{hendrycks2021measuring} datasets as the training datasets and chose five common mathematical reasoning benchmarks to evaluate performance. We conducted our experiments on two models and used GRPO, GSPO, Dr.GRPO  \citep{liu2025understanding}, GRPO with clip-high \citep{yu2025dapo} and GRPO with CLIP-Cov \citep{cui2025entropy} as baselines. The experimental results show that SSPO achieved the highest average score among these five datasets on the 1.5B and 7B models.

\section{Related Works}
\subsection{GRPO reinforcement learning algorithm}
In recent years, reinforcement learning algorithms have achieved great success in the field of training large models, especially the successful application of the GRPO algorithm on DeepSeek-R1 \citep{guo2025deepseek}, which has led to a qualitative leap in inference-related tasks. Our work is based on the improvement of the GRPO algorithm, introducing a sentence-level importance ratio and sentence-level dynamic importance ratio clipping on the basis of the original GRPO algorithm. This addresses the issues of oscillation, unstable training, and entropy collapse that are prone to occur during the training process of the GRPO algorithm
\subsection{Importance ratio}
Importance sampling plays a central role in policy-gradient RL, and is a core component of GRPO-style updates. In GRPO, the importance ratio is computed at the token level. While this fine-grained design directly optimizes each token, it also makes the update highly sensitive to outliers. As a result, the variance and token-level noise can be substantial, and these effects compound with increasing response length, which may eventually trigger unstable learning dynamics or even training collapse.

To address this issue, GSPO computes the importance ratio at the response level, which reduces variance and mitigates the accumulation of token-level noise. However, assigning a single ratio to an entire response introduces a different failure mode under PPO-style clipping: extreme token-level deviations can be “washed out” when aggregated, causing the clipping fraction to approach zero and effectively retaining whole responses even when parts of them are unreliable. This weakens the intended regularization effect of clipping and can lead to unstable policy updates.

We therefore propose SSPO, which computes importance ratios at the subsentence (truncated-sentence) level: tokens within the same subsentence share one importance ratio. This intermediate granularity strikes a balance between GRPO and GSPO—preserving GSPO’s stability benefits while avoiding the tendency to indiscriminately retain entire responses—thereby enabling more reliable and data-efficient learning.

Concurrent work proposed the LPO \citep{li2025every} 10 days before our submission date. LPO proposed Sentence granularity Importance Sampling, which is similar to ours, and validated its effectiveness in a one-trillion-parameter model. However, LPO adopts token-level aggregation, where each token contributes equally to the overall loss. In contrast, we use response-level aggregation, where each response is weighted equally in the loss—consistent with GSPO. Additionally, LPO uses a standard fixed \textit{PPO-Clip} \citep{schulman2017proximal}, while SSPO introduces a novel \textit{adaptive entropy clipping} mechanism (Sec. \ref{sec4.2}), which is designed to dynamically adjust the magnitude of updates and mitigate entropy collapse.

\subsection{Clipping mechanism}
The PPO-CLIP algorithm was proposed in PPO to restrict the confidence interval and improve the stability of reinforcement learning training by limiting the importance ratio. It has also been extended to GRPO. CISPO \citep{chen2025minimax} observed that the gradient of the clipped token does not participate in backpropagation and cannot obtain gradient updates, so it proposed soft clipping gradient. DCPO \citep{yang2025dcpo} uses dynamic clipping to dynamically set different clipping ranges based on the probability of different tokens. Our algorithm introduces the clipping base entropy, taking the entropy of the token as an important factor in calculating the clipping range, which better stabilizes the changes in entropy in reinforcement learning and alleviates the problem of entropy collapse.

\section{Preliminaries}
Notation In this paper, we use $x$ to denote a query and $D$ to denote the query set. Given a response $y$ to a query $x$, $|y|$ denotes the number of tokens in $y$.
\subsection{Group Relative Policy Optimization}  \citet{shao2024deepseekmath} proposed GRPO. The value model was removed by calculating the relative advantage of each response within a set of responses to the same query. GRPO optimizes the following objective:

\begin{equation}
\begin{aligned}
    &\mathscr{J}_{\text{GRPO}}(\theta) = \mathbb{E}_{\substack{x \sim D, \\ \{y_i\} \sim \pi_{\theta_{\text{old}}}(\cdot|x)}} \Bigg[ \frac{1}{G}\sum_{i=1}^{G} \frac{1}{|y_i|} \sum_{t=1}^{|y_i|} \\
    &\min \bigg( w_{i,t}(\theta)\hat{A}_{i,t},\\
    &\text{clip}\Big(w_{i,t}(\theta), 1-\epsilon, 1+\epsilon\Big)\hat{A}_{i,t} \bigg) \Bigg]
\end{aligned}
\end{equation}
Where $G$ is the number of responses generated for each query, $\theta_{old}$ is the old policy, $\epsilon$ is the clipping range of the importance ratios. The importance ratio ${w_{i,t}}\left ({\theta}\right )$ and the advantage ${\hat{A}}_{i,t}$ of token $y_{i,t}$ are:

\begin{equation}
w_{i,t}(\theta)=\frac{\pi_{\theta}(y_{i,t}|x,y_{i,<t})}{\pi_{\theta_{old}}(y_{i,t}|x,y_{i,<t})}
\end{equation}
\begin{equation}
\hat{A}_{i,t}=\hat{A}_i=\frac{r(x,y_i)-\text{mean}(\{r(x,y_i\}_{i=1}^G)}{\text{std}(\{r(x,y_i\}_{i=1}^G)}
\end{equation}
\subsection{Group Sequence Policy Optimization}\citet{zheng2025group} proposed GSPO. GSPO defines the importance ratio based on sequence likelihood and performs response-level clipping. GSPO optimizes the following objective:
\begin{equation}
\begin{split}
&{{\mathscr{J}}_{GSPO}}\left ({\theta}\right ) = {E_{\text{x$\sim $D,}{{\{y_i\}}_{i=1}^G}\text{$\sim $}{\pi_{{\theta_{old}}}}(.|x)}}\Bigg[\frac{1}{G}\sum_{i=1}^{G}\\\min
&\left({s_{i}}\left ({\theta}\right ){{\hat{A}}_{i}},\, \operatorname{clip}\left({s_{i}}\left ({\theta}\right ),1-\epsilon,1+\epsilon\right){{\hat{A}}_{i}}\right)\Bigg]
\end{split}
\end{equation}

The importance ratio ${s_{i}}\left ({\theta}\right )$ and advantage ${\hat{A}}_{i}$ are:
\begin{equation}
\hat{A}_i=\frac{r(x,y_i)-\text{mean}(\{r(x,y_i\}_{i=1}^G)}{\text{std}(\{r(x,y_i\}_{i=1}^G)}
\end{equation}

\begin{equation}
\begin{split}
&s_{i}(\theta)=(\frac{\pi_{\theta}(y_{i}|x)}{\pi_{\theta_{old}}(y_{i}|x)})^{\frac{1}{|y_{i}|}}\\
&=exp({\frac{1}{|y_{i}|}}\sum_{t=1}^{|y_{i}|}log\frac{\pi_{\theta}(y_{i,t}|x,y_i<t)}{\pi_{\theta_{old}}(y_{i,t}|x,y_i<t)})
\end{split}
\end{equation}
\section{Methodology}
\subsection{Subsentence-level importance ratio}
The importance ratio of GRPO is token-level and is easily affected by outliers, leading to unstable training. GSPO designs the calculation of the importance ratio at the response level, and uses length normalization to reduce the policy gradient variance, reduce the impact of extreme tokens, and ensure the stability of training. However, the sequence-level importance ratio calculation method results in a large proportion of clipped tokens, which reduces sample utilization. In order to maintain the stable advantages of GSPO training and increase sample utilization, we propose subsentence-level importance ratio. We segment the samples into multiple fragments by line break or double line break (which we regard as the smallest semantic unit)  and calculate the importance ratio for each fragment. Without considering clips, the optimization objectives are as follows:
\begin{equation}
\begin{split}
&{{\mathscr{J}}_{SSPO}}\left ({\theta}\right )\text{= }{E_{\text{x$\sim $D,}{{\{y_i\}}_{i=1}^G}\text{$\sim $}{\pi_{{\theta_{old}}}}(.|x)}}\\
&\left[\frac{1}{G}\sum_{i=1}^{G}\frac{1}{|y_{i}|}\sum_{j=1}^{N_{seg}(y_{i})}|y_{i,j}|s_{i,j}(\theta){{{\hat{A}}_{i,j}}}\right]
\end{split}
\end{equation}
We adopt group-based advantage estimation:
\begin{equation}
\hat{A}_{i,j}=\hat{A}_i=\frac{r(x,y_i)-\text{mean}(\{r(x,y_i\}_{i=1}^G)}{\text{std}(\{r(x,y_i\}_{i=1}^G)}
\end{equation}
$y_{i,j}$ represents the j-th segment after the i-th sample is semantically segmented. $N_{seg}(y_{i})$ represents the number of segments in $y_{i}$. $L_{i,j,t}$ denote the global index of the $t$-th token in the $j$-th subsentence within the $i$-th sampled response Similarly to GSPO, we define the importance ratio $s_{i,j}(\theta)$ as follows:
\begin{equation}
\begin{split}
&s_{i,j}(\theta)=(\frac{\pi_{\theta}(y_{i,j}|x)}{\pi_{\theta_{old}}(y_{i,j}|x)})^{\frac{1}{|y_{i,j}|}}\\
&=exp({\frac{1}{|y_{i,j}|}}\sum_{t=1}^{|y_{i,j}|}log\frac{\pi_{\theta}(y_{i,j,t}|x,y_{i,}<L_{i,j,t})}{\pi_{\theta_{old}}(y_{i,j,t}|x,y_{i,}<L_{i,j,t})})
\end{split}
\end{equation}
The gradient objective of SSPO is:
\begin{equation}
\begin{split}
\nabla_{\theta}\mathscr{J}_{\text{SSPO}}(\theta)
= \mathbb{E}_{x \sim D,\ \{y_i\}_{i=1}^{G} \sim \pi_{\theta_{\text{old}}}(\cdot \mid x)} \Bigg[ \\
\frac{1}{G}\sum_{i=1}^{G}\frac{1}{|y_i|}\sum_{j=1}^{N_{\text{seg}}(y_i)} |y_{i,j}|
\left( \frac{\pi_{\theta}(y_{i,j}\mid x)}{\pi_{\theta_{\text{old}}}(y_{i,j}\mid x)} \right)^{\frac{1}{|y_{i,j}|}} \hat{A}_i \cdot \\
\frac{1}{|y_{i,j}|}\sum_{t=1}^{|y_{i,j}|} \nabla_{\theta}\log \pi_{\theta}(y_{i,j,t}\mid x, y_{i,j,<t})
\Bigg]
\end{split}
\end{equation}
Detailed derivations of the gradient objective of SSPO can be found in \ref{sec:appendix}. For comparison, GRPO's gradient objective is as follows:

\begin{equation}
\begin{split}
\nabla_{\theta}\mathscr{J}_{\text{GRPO}}(\theta)
= \mathbb{E}_{x \sim D,\ \{y_i\}_{i=1}^{G} \sim \pi_{\theta_{\text{old}}}(\cdot \mid x)} \Bigg[ \\
\frac{1}{G}\sum_{i=1}^{G}\hat{A}_i \cdot \frac{1}{|y_i|}\sum_{t=1}^{|y_i|}
\left( \frac{\pi_{\theta}(y_{i,t} \mid x, y_{i,<t})}{\pi_{\theta_{\text{old}}}(y_{i,t} \mid x, y_{i,<t})} \right) \\
\nabla_{\theta}\log \pi_{\theta}(y_{i,t} \mid x, y_{i,<t})
\Bigg]
\end{split}
\end{equation}
The weight of each token gradient of GRPO is different, which causes the problem of high variance of the policy gradient. The gradient target of SSPO is ultimately similar to GSPO. The gradient weights of all tokens in a fragment are equal, eliminating noise interference between tokens, reducing the variance of the policy gradient, and making training more stable.
\subsection{Adaptive Entropy Clipping Mechanism}
\label{sec4.2}
In reinforcement learning, clipping is used to control the magnitude of policy updates to ensure that changes in the old and new policies are within a certain range and enhance the stability of training. However, the clipping mechanism will inhibit the probability growth of low-probability tokens and, at the same time, tend to maintain the original high-probability tokens, resulting in rapid model fitting and entropy collapse. In order to alleviate the phenomenon of entropy collapse and encourage exploration, we propose a dynamic entropy clipping method.
For each token $y_t$ of the output $y$, when the vocabulary size is $V$ pairs, the entropy of the current token under the old strategy is:
\begin{equation}
{{\mathcal{H}}_{t}}=-\sum_{v \in V}^{}{\pi_{\theta_{old}}(v|x,y_{<t})log\pi_{\theta_{old}}}(v|x,y_{<t})
\end{equation}
Since our importance ratio is in the semantic segment dimension, we define the entropy of the semantic segment dimension as follows:
\begin{equation}
{{\mathcal{H}}_{i,j}}=\frac{1}{|y_{i,j}|} \sum_{t=1}^{|y_{i,j}|}{\mathcal{H}}_{t}
\end{equation}
We define the dynamic entropy clipping method as follows:
\begin{equation}
b_{high}=min(1+\alpha+{\mathcal{H}_{i,j}},\gamma)
\end{equation}
\begin{equation}
b_{low}=max(0,min(1-{\mathcal{H}_{i,j}},\beta))
\end{equation}
When the entropy of the subsentence is relatively large, we expand the clip interval range, allowing tokens with high entropy to participate more in the gradient update. This prevents them from being clipped and increases the update intensity. When the entropy of the subsentence is relatively small, it means that the model has learned these tokens well. Therefore, we reduce the clipping interval bound to limit the update intensity of these tokens. $b_{low}$ and $b_{high}$ represent the lower bound and upper bound of clipping range. $\alpha$, $\beta$ and $\gamma$ are three hyperparameters. 
Through semantic segmentation and adaptive entropy clipping, our final objective is as follows:
\begin{equation}
\begin{split}
\mathscr{J}_{\text{SSPO}}(\theta)
= \mathbb{E}_{x \sim D,\ \{y_i\}_{i=1}^{G} \sim \pi_{\theta_{\text{old}}}(\cdot \mid x)} \Bigg[ \\
\frac{1}{G}\sum_{i=1}^{G}\frac{1}{|y_{i}|}\sum_{j=1}^{N_{\text{seg}}(y_{i})}|y_{i,j}|
\min\Big( s_{i,j}(\theta)\hat{A}_{i}, \\
\text{clip}\big(s_{i,j}(\theta),\, b_{\text{low}},\, b_{\text{high}}\big)\hat{A}_{i,j} \Big)
\Bigg]
\end{split}
\end{equation}

\section{Experimental Setup}
\subsection{Model} We experimented with two mathematical models Qwen2.5-Math-1.5B and Qwen2.5-Math-7B \citep{yang2024qwen2}.
\subsection{Training} Following Dr.GRPO, we choose MATH Level 3-Level 5 as our training datasets, which contain 8523 mathematical problems. For each question, we sample 8 rollouts and set the model’s response length at 3000 tokens and set the batch size to 128. All problems were processed using the Qwen-Math template for both training and evaluation. In all experiments of the RL algorithms, we use the veRL \citep{sheng2024hybridflow} framework to support our experiments. In our adaptive entropy clipping mechanism, we use settings $\alpha$=0, $\beta$=0.8, $\gamma$=1.75 on Qwen2.5-Math-1.5B and use settings $\alpha$=0.1, $\beta$=0.8, $\gamma$=2.2 on Qwen2.5-Math-7B.

\subsection{Baseline Settings}We use GRPO, GSPO, GRPO with clip-high and GRPO with CLIP-Cov as baselines

\subsection{Evaluation}We evaluated our models on five widely used mathematical reasoning benchmarks: MATH-500 \citep{hendrycks2021measuring}, a subset of 500 problems from MATH datasets including algebra, geometry, and number theory; AMC \citep{li2024numinamath} consists of 83 intermediate-difficulty multiple-choice problems; AIME24 \citep{li2024numinamath} contains 30 olympiad problems from American Invitational Mathematics Examination 2024; Minerva \citep{lewkowycz2022solving} comprises 272 graduate-level multi-step reasoning problems; Olympiad Bench \citep{he2024olympiadbench} owns 675 high-difficulty Olympiad questions.
\section{Result and Analysis}
\subsection{Main Results}

\begin{table*}[t]
\centering
\caption{Comparison of methods on Qwen2.5-Math-1.5B and Qwen2.5-Math-7B on five mathematical datasets. Method's results marked with * mean directly using the experimental results reported by the paper of this method}
\label{tab:main}
\setlength{\tabcolsep}{3pt} 
\renewcommand{\arraystretch}{1.0} 

\begin{tabular}{l
                S[table-format=2.2]
                S[table-format=2.2]
                S[table-format=2.2]
                S[table-format=2.2]
                S[table-format=2.2]
                S[table-format=2.2]}
\toprule
\textbf{Method} &
\textbf{AIME24} & \textbf{AMC23} & \textbf{MATH} & \textbf{MIN.} & \textbf{OLY.} & \textbf{Avg} \\
\midrule

\multicolumn{7}{c}{\textbf{Qwen2.5-Math-1.5B}} \\
\midrule
GRPO                    & 16.67 & 54.20 & 72.60 & 32.35 & 39.67 & 43.01 \\
GSPO                    & 20.00 & 51.49 & 74.60 & 34.56 & 41.16 & 44.42 \\
Dr.\ GRPO*               & 20.00 & 53.00 & 74.20 & 25.70 & 37.60 & 42.10 \\
GRPO($\epsilon_{\text{high}}=0.28$)                  & 13.30 & 53.01 & 73.20 & 33.09 & 37.89 & 42.10 \\
CLIP-Cov                    & 20.00 & 56.63 & \textbf{78.0} & 35.66 & 40.56 & 46.17 \\
SSPO                    & \textbf{23.33} & \textbf{55.42} & 76.00 & \textbf{36.42} & \textbf{42.50} & \textbf{46.72} \\
\midrule

\multicolumn{7}{c}{\textbf{Qwen2.5-Math-7B}} \\
\midrule
GRPO                    & 33.30 & 67.47 & 79.00 & 40.07 & 45.91 & 53.15 \\
GSPO                    & 33.30 & 65.06 & 80.80 & 42.28 & 47.10 & 53.75 \\
Dr.\ GRPO*               & \textbf{43.30} & 62.70 & 80.00 & 30.10 & 41.00 & 51.40 \\
GRPO($\epsilon_{\text{high}}=0.28$)                    & 33.30 & 68.67 & 82.2 & \textbf{44.49} & \textbf{49.18} & 55.57 \\
CLIP-Cov                    & 36.67 & 67.47 & \textbf{84.60} & 41.91 & 46.95 & 55.52 \\
SSPO                    & 40.0 & \textbf{69.88} & 81.20 & 41.54 & 47.40 & \textbf{56.00} \\
\bottomrule
\end{tabular}
\end{table*}

Table \ref{tab:main} summarizes the performance of various RL optimization methods, including GRPO, Dr.GRPO, GSPO, GRPO with clip-high and GRPO with CLIP-Cov, and our proposed SSPO, on five mathematical reasoning benchmarks: MATH500, AMC23, AIME24, Minerva, and Olympiad. We evaluate these datasets basically on greedy decoding(Avg@1) to intuitively present the effectiveness of our method. We evaluate all approaches in five datasets every 10 steps on both 1.5B and 7B models, and choose the best score as our final score.  Among the Qwen2.5-Math-1.5B and Qwen2.5-Math-7B models, SSPO offers superior performance relative to other methods. Table~\ref{tab:main} reports the main results on five math benchmarks (AIME24, AMC23, MATH, MIN., and OLY.) under two backbone sizes (Qwen2.5-Math-1.5B and Qwen2.5-Math-7B). Overall, SSPO delivers the most consistent gains across both model scales, achieving the best average performance in each setting.

On Qwen2.5-Math-1.5B, SSPO attains the highest overall average score (46.72), outperforming strong PPO-style baselines such as GRPO (43.01) and GSPO (44.42), and also surpassing the competitive CLIP-Cov variant (46.17). Notably, SSPO achieves the best performance on four of the five datasets (AIME24, AMC23, MIN., and OLY.), indicating robust improvements that generalize beyond a single benchmark. Even on MATH, where CLIP-Cov is strongest (78.0), SSPO remains competitive (76.0), preserving strong aggregate performance.

On the larger Qwen2.5-Math-7B model, SSPO again yields the best average (56.00), exceeding GRPO (53.15), GSPO (53.75), and CLIP-Cov (55.52). While some baselines win on individual datasets (e.g., Dr. GRPO* on AIME24 and CLIP-Cov on MATH), SSPO provides the most stable cross-dataset improvements, ranking first on AMC23 and maintaining competitive results on the remaining benchmarks. These results suggest that SSPO scales favorably with model capacity and improves overall generalization, rather than trading off gains on one dataset for regressions on others.
\subsection{Comparative experiments}
In this subsection, we conduct extensive ablation studies by progressively augmenting a naive GSPO baseline with subsentence-level importance ratio and adaptive entropy clipping, thereby validating the effectiveness of both components. In addition, we perform comprehensive comparative experiments to identify the optimal sentence segmentation strategy and the best hyperparameter configuration for adaptive entropy clipping.
\subsubsection{Ablation study} Table~\ref{tab2} reports an ablation study to quantify the impact of adaptive entropy clipping in SSPO. We compare GSPO, SSPO without entropy clip (i.e., fixed clipping bounds), and the full SSPO. Entropy clip dynamically adjusts the clipping interval based on the average token entropy of each subsentence observed during training, with the goal of preserving exploration while keeping policy updates stable—tightening constraints when the policy becomes over-confident and relaxing them when uncertainty is higher.

Across both model scales, entropy clip yields consistent and non-trivial gains over SSPO without this mechanism. For Qwen2.5-Math-1.5B, SSPO improves the average score from 45.54 (w/o entropy clip) to 46.72, and achieves the best results on all five benchmarks, with particularly large improvements on MIN. (32.72→36.40) and OLY. (39.52→42.50), which are more sensitive to exploration–exploitation trade-offs. For Qwen2.5-Math-7B, entropy clipping further raises the average from 53.99 to 54.85, with clear gains on AIME24 (33.30→36.67) and AMC23 (65.06→66.27). These results indicate that adaptively entropy clipping is a key driver of SSPO’s robustness and cross-dataset generalization.

\begin{table}[t]
\centering
\caption{Main results of progressive techniques applied to SSPO. +SIR represent applying subsentence-level importance ratio. +EC represent apply both subsentence-level importance ratio and adaptive entropy clipping. }
\label{tab2}
\setlength{\tabcolsep}{3.2pt} 
\renewcommand{\arraystretch}{1.05}
\small 
\begin{tabular}{@{}lcccccc@{}}
\toprule
\textbf{Method} & \textbf{AIME24} & \textbf{AMC} & \textbf{MATH} & \textbf{MIN.} & \textbf{OLY.} & \textbf{Avg} \\
\midrule
\multicolumn{7}{c}{\textit{Qwen2.5-Math-1.5B}} \\
\midrule
GSPO & 20.00 & 51.49 & 74.6  & 34.56 & 41.16 & 44.42 \\
\makecell[l]{+ SIR} & 20.00 & 56.63 & 75.00 & 34.19 & 41.90 & 45.54 \\
+ EC & \textbf{23.33} & \textbf{57.83} & \textbf{75.40} & \textbf{36.40} & \textbf{42.50} & \textbf{46.72} \\
\midrule
\multicolumn{7}{c}{\textit{Qwen2.5-Math-7B}} \\
\midrule
GSPO & 33.30 & 65.06 & 80.8  & \textbf{42.28} & 47.1  & 53.75 \\
\makecell[l]{+ SIR} & 33.30 & 65.06 & \textbf{81.61} & 42.28 & \textbf{47.70} & 53.99 \\
+ EC & \textbf{40.00} & \textbf{69.88} & 81.2 & 41.54 & 47.40 & \textbf{56.0} \\
\bottomrule
\end{tabular}
\end{table}

\begin{table}[t]
\centering
\caption{Comparison of three sentence segmentation strategies: period (.), line break (\textbackslash n), and double line break (\textbackslash n\textbackslash n).}
\label{tab3}
\setlength{\tabcolsep}{3.2pt}
\renewcommand{\arraystretch}{1.05}
\small 
\begin{tabular}{@{}lcccccc@{}}
\toprule
\textbf{Seg.} & \textbf{AIME24} & \textbf{AMC} & \textbf{MATH} & \textbf{MIN.} & \textbf{OLY.} & \textbf{Avg} \\
\midrule
\multicolumn{7}{c}{\textit{Qwen2.5-Math-1.5B}} \\
\midrule
(.) & 20.00 & 53.01 & \textbf{75.00} & 30.88 & 39.67 & 43.71 \\
(\textbackslash n) & \textbf{23.33} & \textbf{56.63} & 74.20 & 32.72 & 39.52 & 45.27 \\
(\textbackslash n\textbackslash n) & 20.00 & \textbf{56.63} & \textbf{75.00} & \textbf{34.19} & \textbf{41.90} & \textbf{45.54} \\
\bottomrule
\end{tabular}
\end{table}

\subsubsection{Segmentation Strategy for SSPO}
Table~\ref{tab3} compares SSPO trained with three segmentation rules—splitting by period (“.”), by a single line break (\verb|\n|), and by double line breaks (\verb|\n|\verb|\n|)—and finds that double line breaks yield the best overall performance (Avg 45.54 vs. 45.27 for \verb|\n| and 43.71 for periods), with particularly clear gains on MIN. and OLY. We hypothesize that \verb|\n|\verb|\n| better matches the natural structure of many reasoning traces, where blank lines often mark the end of a coherent semantic step (e.g., separating intermediate conclusions or sub-cases), enabling SSPO to optimize over more meaningful segments and thus improving generalization across datasets. Therefore, we use double line breaks as the delimiter for sentence segmentation. 

\subsubsection{Hyperparameters in entropy clipping mechanism}
In Eqs. (14) (15), $\alpha$, $\beta$, and $\gamma$ control the entropy-adaptive clipping interval. $\alpha$ provides an additive slack to the upper bound, enlarging the allowable update magnitude for high-entropy (uncertain) subsentences and thus encouraging exploration. $\gamma$ acts as a hard cap on the upper bound to prevent overly aggressive ratio expansion and stabilize optimization. $\beta$ caps the entropy-dependent lower bound, tightening the clipping range for low-entropy (confident) subsentences to limit unnecessary updates and mitigate collapse.We determine the initial hyperparameter search ranges by analyzing the empirical distribution of subsentences entropy $H_{i,j}$ from a short pilot run, and by referencing the standard PPO-Clip setting with $\epsilon$ = 0.2. Since $b_{\text{high}}$ is governed by $1+\alpha+H_{i,j}$ and capped by $\gamma$, we set $\gamma$ around the typical values of $1+\alpha+H_{i,j}$ (median to high-percentile with a small margin). Similarly, $\beta$ is chosen to match the scale of $1-H_{i,j}$, ensuring the lower bound is neither trivially saturated nor inactive. We conducted exploratory experiments within the initial hyperparameter search ranges on Qwen2.5-Math-1.5B(see Table \ref{tab4})  and Qwen2.5-Math-7B(see Table \ref{tab5}) and ultimately identified the most suitable values for the three hyperparameters for each models.

\begin{table}[t]
\centering
\caption{We investigate the optimal values for each parameter and explore the hyperparameter ranges by fixing two parameters at a time and varying the remaining one. Table 4 represents the exploratory experiments on Qwen2.5-Math-1.5B. }
\label{tab4}
\setlength{\tabcolsep}{3.2pt} 
\renewcommand{\arraystretch}{1.05}
\small 
\begin{tabular}{@{}lcccccc@{}}
\toprule
\textbf{Method} & \textbf{AIME24} & \textbf{AMC} & \textbf{MATH} & \textbf{MIN.} & \textbf{OLY.} & \textbf{Avg} \\
\midrule
\multicolumn{7}{c}{\textit{$\alpha=0.0, \beta=0.8$}} \\
\midrule
$\gamma$=1.5 & 20.00 & 54.22 & 74.80  & 35.66 & 41.31 & 45.20 \\
\makecell[l]{$\gamma$=1.6} & 20.00 & 55.42 & 74.20 & 34.56 & 39.23 & 44.68 \\
$\gamma$=1.7 & 16.67 & 57.83 & 74.60 & 33.46 & 40.42 & 44.60 \\
$\gamma$=1.75 & 23.33 & 55.42 & 76.00 & 36.40 & 42.50 & \textbf{46.72} \\
$\gamma$=1.8 & 23.33 & 54.22 & 75.20 & 34.19 & 40.71 & 45.52 \\
\midrule
\multicolumn{7}{c}{\textit{$\beta=0.8, \gamma=1.75$}} \\
\midrule
$\alpha$=0 & 23.30 & 55.42 & 76.00  & 36.40 & 42.50 & \textbf{46.72} \\
\makecell[l]{$\alpha$=0.05} & 20.00 & 53.01 & 75.40 & 35.66 & 40.71 & 45.00 \\
$\alpha$=0.1 & 23.33 & 55.42 & 74.40 & 34.93 & 40.86 & 45.78 \\
$\alpha$=0.15 & 20.00 & 53.01 & 75.60 & 34.93 & 40.27 & 44.76 \\
$\alpha$=0.2 & 23.33 & 53.01 & 75.60 & 36.40 & 40.56 & 45.77 \\
\midrule
\multicolumn{7}{c}{\textit{$\alpha=0, \gamma=1.75$}} \\
\midrule
$\beta$=0.7 & 23.33 & 53.01 & 75.00  & 32.72 & 40.12 & 44.84 \\
\makecell[l]{$\beta$=0.8} & 23.33 & 55.42 & 76.00 & 36.40 & 42.50 & \textbf{46.72} \\
$\beta$=0.9 & 23.33 & 54.22 & 74.00 & 34.19 & 39.67 & 45.08 \\
\bottomrule
\end{tabular}
\end{table}

\begin{table}[t]
\centering
\caption{Table 5 represents the exploratory experiments on Qwen2.5-Math-7B. }
\label{tab5}
\setlength{\tabcolsep}{3.2pt} 
\renewcommand{\arraystretch}{1.05}
\small 
\begin{tabular}{@{}lcccccc@{}}
\toprule
\textbf{Method} & \textbf{AIME24} & \textbf{AMC} & \textbf{MATH} & \textbf{MIN.} & \textbf{OLY.} & \textbf{Avg} \\
\midrule
\multicolumn{7}{c}{\textit{$\alpha=0, \beta=0.8$}} \\
\midrule
$\gamma$=2.0 & 30.00 & 67.47 & 81.8  & 41.91 & 47.70 & 53.78 \\
\makecell[l]{$\gamma$=2.1} & 33.33 & 65.06 & 80.00 & 42.28 & 46.95 & 53.52 \\
$\gamma$=2.2 & 36.67 & 67.47 & 81.20 & 43.75 & 47.10 & \textbf{55.24} \\
\midrule
\multicolumn{7}{c}{\textit{$\beta=0.8, \gamma=2.2$}} \\
\midrule
$\alpha$=0 & 36.67 & 67.47 & 81.20 & 43.75 & 47.10 & 55.24 \\
\makecell[l]{$\alpha$=0.1} & 33.33 & 69.88 & 81.20 & 42.65 & 47.40 & 54.89 \\
$\alpha$=0.15 & 40.00 & 69.88 & 81.20 & 41.54 & 47.40 & \textbf{56.00} \\
$\alpha$=0.2 & 36.67 & 67.47 & 81.20 & 41.91 & 48.59 & 55.12 \\
\midrule
\multicolumn{7}{c}{\textit{$\alpha=0.15, \gamma=2.2$}} \\
\midrule
$\beta$=0.7 & 33.33 & 63.86 & 80.40  & 43.01 & 47.25 & 53.56 \\
\makecell[l]{$\beta$=0.8} & 40.00 & 69.88 & 81.20 & 41.54 & 47.40 & \textbf{56.00} \\
$\beta$=0.9 & 30.00 & 69.88 & 81.00 & 41.54 & 49.18 & 54.32 \\
\bottomrule
\end{tabular}
\end{table}

\subsection{Observational study} In this subsection, we validate the advantages of our proposed SSPO algorithm over GRPO and GSPO in terms of training stability and the model’s exploration capability by examining several interesting empirical phenomena.
\subsubsection{Entropy decrease on different RL methods}
Figure \ref{fig:entropy_both} illustrate that across both model scales, SSPO shows a slower and smoother entropy decay, keeping the policy at a relatively higher entropy level for longer during training. This helps preserve exploration and mitigates premature entropy collapse observed in GRPO/GSPO, leading to more stable optimization dynamics.
\subsubsection{Fractions of clipped tokens}
Figure \ref{fig:clipfrac} reports the fraction of tokens whose importance ratios are clipped by the PPO-style surrogate objective throughout RL training for GSPO, GRPO, and SSPO (with entropy clipping). GRPO consistently exhibits the largest clipped-token fraction, indicating frequent violations of the trust-region constraint and more aggressive effective updates. SSPO clips substantially fewer tokens, with intermittent spikes, suggesting improved stability and reduced mismatch between the updated and reference policies. In contrast, GSPO remains at (or extremely close to) zero across all training steps, implying that its PPO-clipping mechanism is rarely activated and that policy ratio updates almost always stay within the clipping range. We observe that, in GSPO, few tokens are clipped during training, meaning that even extreme tokens in the sampled data are retained.
\subsubsection{Training rewards of different RL methods}
Figure~\ref{fig:reward} compares the training reward trajectories of GRPO, GSPO, and SSPO w. entropy clip over training steps. SSPO exhibits a sustained and nearly monotonic increase in training reward, rising steadily from the early phase and continuing to improve throughout training while remaining at a high level thereafter. In contrast, GRPO improves initially but then deteriorates sharply in the mid-to-late stage, eventually collapsing to near-zero reward. GSPO shows a slower upward trend overall, but with noticeably larger oscillations and several abrupt drops in the late stage. Overall, SSPO delivers the most consistently improving reward curve across the entire training horizon.

\section{Conclusion}
We propose SSPO, a stable variant of GRPO. By applying subsentence-level importance ratio and adaptive entropy clipping, SSPO not only allows models to improve the performance of mathematical benchmarks, but also alleviates the entropy collapse, encouraging models to explore and learn. Our experiments indicate that SSPO outperforms GRPO, Dr.GRPO, GSPO, GRPO with clip-high and GRPO with CLIP-Cov in both 1.5B and 7B models under five mathematical datasets. This work makes a contribution to developing more reliable and scalable reinforcement learning for LLMs, and future work will focus on extending SSPO to other domains, such as programming and semantic reasoning.

\section*{Limitations}
Our study has several limitations. First, due to constrained training resources, we primarily evaluate SSPO on mathematical reasoning models and benchmarks; it remains unclear how well the method generalizes to other domains, model families, or training setups. Second, our current response segmentation relies on simple heuristic delimiters (e.g., line breaks). More semantically grounded segmentation—such as using dedicated discourse/semantic segmentation models—may yield more coherent subsentences and further improve the reliability of subsentence-level updates. Third, although we perform exploratory tuning for $\alpha$, $\beta$, and $\gamma$, a more rigorous hyperparameter search protocol (e.g., principled range selection, interaction-aware search, and statistically grounded selection criteria) is needed to better characterize the sensitivity of entropy-adaptive clipping and to ensure reproducible optimal settings.

\section*{Ethics Statement}
This work introduces SSPO to enhance the mathematical reasoning capabilities of Large Language Models using exclusively publicly available datasets that contain no personally identifiable information. While reinforcement learning requires computational resources, our approach aims to improve training stability and data efficiency, potentially reducing the long-term environmental footprint associated with model training. We do not foresee immediate negative societal impacts from advancing mathematical reasoning, though we acknowledge the inherent biases present in the pre-trained base models and remain committed to the responsible development of AI.

\bibliography{custom}

\appendix
\onecolumn

\section{Appendix}
\subsection{The gradient of SSPO}
\label{sec:appendix}

The gradient objective of SSPO is as follows:
\begin{equation}
\triangledown_{\theta}{{\mathscr{J}}_{SSPO}}\left ({\theta}\right )\text{= }{\triangledown_{\theta}E_{\text{x$\sim $D,}{{\{y_i\}}_{i=1}^G}\text{$\sim $}{\pi_{{\theta_{old}}}}(.|x)}}\left[\frac{1}{G}\sum_{i=1}^{G}\frac{1}{|y_{i}|}\sum_{j=1}^{N_{seg}(y_{i})}|y_{i,j}|s_{i,j}(\theta){{{\hat{A}}_{i,j}}}\right]
\end{equation}

\begin{equation}
=E_{\text{x$\sim $D,}{{\{y_i\}}_{i=1}^G}\text{$\sim $}{\pi_{{\theta_{old}}}}(.|x)}\left[\frac{1}{G}\sum_{i=1}^{G}\frac{1}{|y_{i}|}\sum_{j=1}^{N_{seg}(y_{i})}|y_{i,j}|s_{i,j}(\theta){{\hat{A}}_{i,j}}\triangledown_{\theta}logs_{i,j}(\theta) \right]
\end{equation}

\begin{multline}
=E_{x \sim D,\{y_{i}\}_{i=1}^{G} \sim \pi_{\theta_{old}}(\cdot | x)} \\
\left[ \frac{1}{G} \sum_{i=1}^{G} \frac{1}{|y_{i}|}\sum_{j=1}^{N_{seg}(y_{i})} |y_{i,j}|\left( \frac{\pi_{\theta}(y_{i,j} | x)}{\pi_{\theta_{old}}(y_{i,j} | x)} \right)^{\frac{1}{|y_{i,j}|}} {{\hat{A}}_{i,j}} \cdot \frac{1}{|y_{i,j}|} \sum_{t=1}^{|y_{i,j}|} \nabla_{\theta} \log \pi_{\theta}(y_{i,j,t} | x, y_{i,j,<t}) \right]
\end{multline}

\begin{multline}
=E_{x \sim D,\{y_{i}\}_{i=1}^{G} \sim \pi_{\theta_{old}}(\cdot | x)} \\
\left[ \frac{1}{G} \sum_{i=1}^{G} \frac{1}{|y_{i}|}\sum_{j=1}^{N_{seg}(y_{i})} |y_{i,j}|\left( \frac{\pi_{\theta}(y_{i,j} | x)}{\pi_{\theta_{old}}(y_{i,j} | x)} \right)^{\frac{1}{|y_{i,j}|}} {{\hat{A}}_{i}} \cdot \frac{1}{|y_{i,j}|} \sum_{t=1}^{|y_{i,j}|} \nabla_{\theta} \log \pi_{\theta}(y_{i,j,t} | x, y_{i,j,<t}) \right]
\end{multline}

\subsection{Figures of training entropy}
\begin{figure}[!htbp]
    \centering
    \begin{subfigure}[t]{0.39\textwidth}
        \centering
        \includegraphics[width=\linewidth]{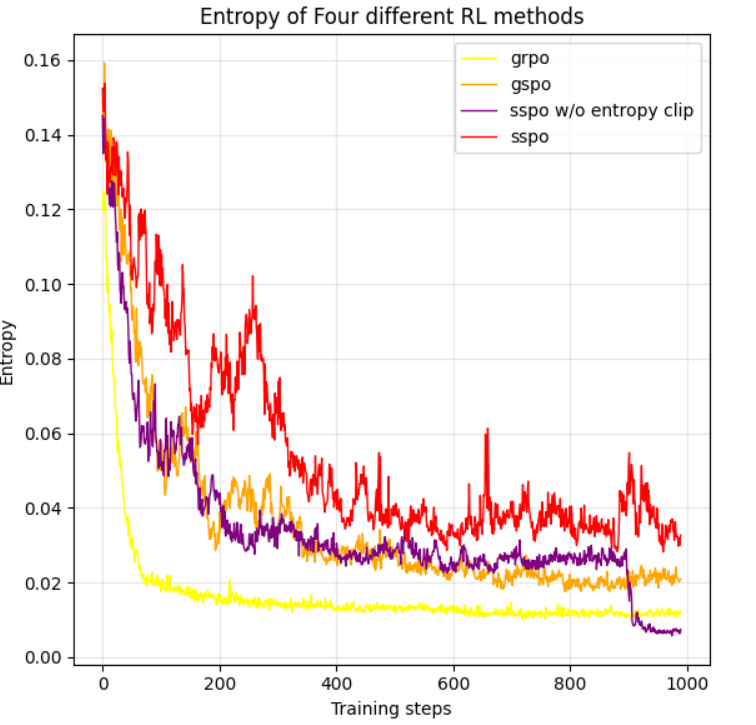}
        \caption{7B model.}
        \label{fig:entropy_7b}
    \end{subfigure}\hspace{2mm}
    \begin{subfigure}[t]{0.38\textwidth}
        \centering
        \includegraphics[width=\linewidth]{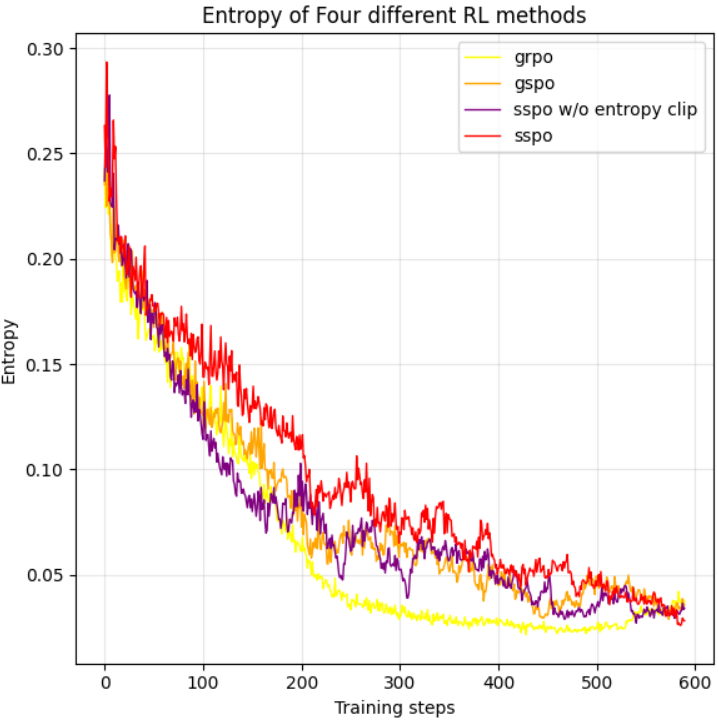}
        \caption{1.5B model.}
        \label{fig:entropy_15b}
    \end{subfigure}
    \caption{Training entropy of SSPO, SSPO (w/o entropy clip), GSPO, and GRPO across model scales.}
    \label{fig:entropy_both}
\end{figure}
\clearpage 

\subsection{Figures of training reward and clipping fraction}
\begin{figure}[!htbp]
    \centering
    \begin{subfigure}[t]{0.37\textwidth}
        \centering
        \includegraphics[width=\linewidth]{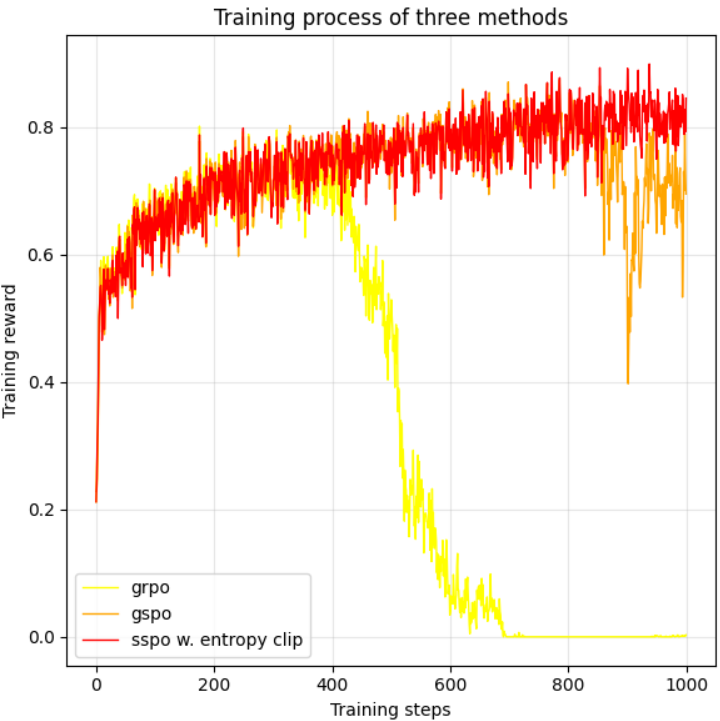}
        \caption{The training reward of 1.5B model using SSPO, GSPO and GRPO}
        \label{fig:reward}
    \end{subfigure}\hspace{2.1mm}
    \begin{subfigure}[t]{0.4\textwidth}
        \centering
        \includegraphics[width=\linewidth]{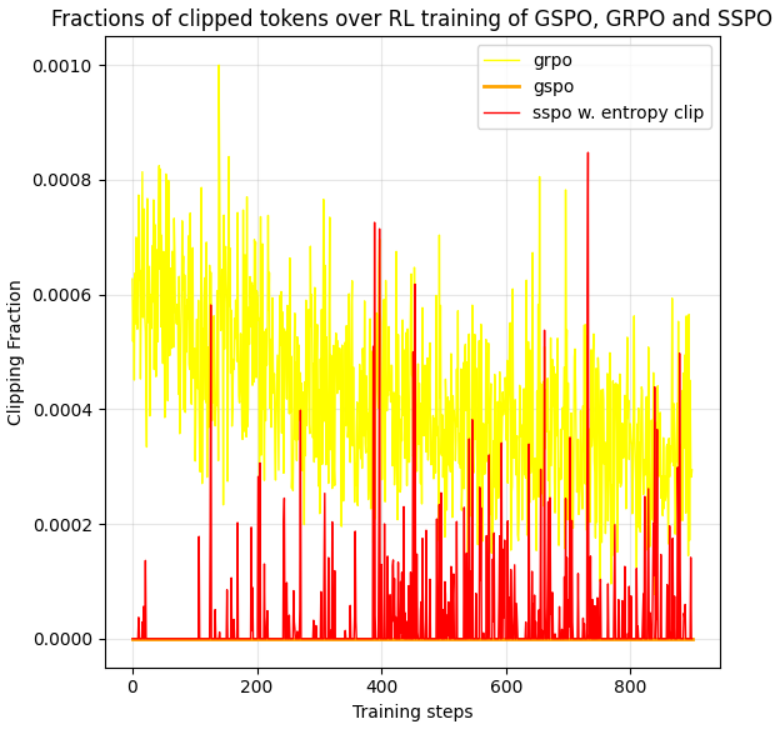}
        \caption{Clipping Fractions of three RL methods}
        \label{fig:clipfrac}
    \end{subfigure}
    \caption{Figure (a) illustrates the training reward of the three algorithms as a function of training steps. Figure (b) reports fractions of clipped tokens as training progresses over three methods.}
    \label{fig:reward_clip}
\end{figure}
\end{document}